\documentclass[conference]{IEEEtran}
\IEEEoverridecommandlockouts
\usepackage{cite}
\usepackage{amsmath,amssymb,amsfonts}
\usepackage{algorithmic}
\usepackage{graphicx}
\usepackage{textcomp}
\usepackage{xcolor}
\def\BibTeX{{\rm B\kern-.05em{\sc i\kern-.025em b}\kern-.08em
    T\kern-.1667em\lower.7ex\hbox{E}\kern-.125emX}}
\begin{document}

\title{Automatically Classifying Emotions based on Text: A Comparative Exploration of Different Datasets}

\author{\IEEEauthorblockN{Anna Koufakou,  Jairo Garciga, Adam Paul, Joseph Morelli and Christopher Frank}
\IEEEauthorblockA{\textit{Department of Computing and Software Engineering, Florida Gulf Coast University, Fort Myers, Florida} \\
Email: akoufakou@fgcu.edu}
}

\maketitle

\begin{abstract}
Emotion Classification based on text is a task with many applications which has received growing interest in recent years. This paper presents a preliminary study with the goal to help researchers and practitioners gain insight into relatively new datasets as well as emotion classification in general. We focus on three datasets that were recently presented in the related literature, and we explore the performance of traditional as well as state-of-the-art deep learning models in the presence of different characteristics in the data. We also explore the use of data augmentation in order to improve performance. Our experimental work shows that state-of-the-art models such as RoBERTa perform the best for all cases. We also provide observations and discussion that highlight the complexity of emotion classification in these datasets and test out the applicability of the models to actual social media posts we collected and labeled.  
\end{abstract}


\section{Introduction}

Recognizing emotion automatically from text is of great interest in many applications from emotion-aware recommender systems to intelligent chatbots  to suicide prevention. Emotion detection or classification is quite different from sentiment analysis, a popular term which has attracted a lot of research in earlier years. Sentiment analysis refers to discovering if the content of a text (opinion, tweet, essay) is positive or negative. This is usually a binary problem (for example, positive vs negative reviews of movies or products bought online) though it could extend to more categories (e.g. very positive or very negative or neutral). Emotion prediction or detection on the other hand deals with recognizing specific emotions in the text, such as anger, sadness, or joy. Emotions are complex and many times not as clear to humans, so automatic detection is a challenging task. Building datasets for this task is also difficult as there are different taxonomies of emotions, e.g. Ekman \cite{ekman1992argument} or Plutchik \cite{plutchik1984emotions}; additionally, assigning a single emotion label to text can be highly subjective.

One of the earliest examples of emotion labeled data is in SemEval 2007 \cite{strapparava-mihalcea-2007-semeval}. More recent work in emotion related datasets include tweets (SemEval-2018 Task 1) \cite{mohammad-etal-2018-semeval}, conversations \cite{li-etal-2017-dailydialog}, and movie subtitles \cite{ohman-etal-2018-creating}, as examples. The authors in \cite{bostan-klinger-2018-analysis} unified 14 popular emotion classification corpora
under one framework. As examples of using deep learning for emotion classification, the authors in \cite{rajabi2020multi} performed multi-label classification on the SemEval2018 Task 1 dataset using a multi-channel, multi-filter CNN-BiLSTM. Various BERT-like models were explored in text-based emotion recognition in \cite{adoma2020comparative}. Recent reviews on text-based emotion detection include \cite{alswaidan2020survey,acheampong2021transformer,nandwani2021review}.

In this paper, we presented a preliminary study of three different datasets related to emotion classification: one based on a UK survey related to COVID-19 \cite{kleinberg2020measuring}; another containing essays written after reading news articles \cite{tafreshi2021wassa}; and one extracted from social media comments \cite{demszky2020goemotions} (we describe the dataset in detail in Section \ref{sec-data}). We explored various traditional and state-of-the-art models and present our findings and observations. Our goal is to show the different focus in each dataset and the complexity of recognizing emotion from different datasets. We also attempted to test the applicability of the models on a few example social media posts we collected and labeled. This work aims to help the researcher and practitioner interested in this field by exploring data with different characteristics and perspectives. Our study is preliminary in that our plan is to include more datasets and models as well as incorporate other features from the data in the classification (e.g. demographics or emotion intensity of words).

The organization of this paper is as follows: Section \ref{sec-data} describes and compares the datasets we used in this work. Section \ref{sec-method} summarizes the models and set up of our experiments, followed by Section \ref{sec-results} which presents the results and observations. Finally, Section \ref{sec-conclusion} includes concluding remarks and future research directions.

\section{Datasets}
\label{sec-data}
In this paper, we experimented with three datasets described below\footnote{Datasets and code available by request}. All datasets are in English and were prepared and presented in 2020 or later. There are several existing datasets that were presented before 2020, e.g. from the SemEval-2018 Task 1) \cite{mohammad-etal-2018-semeval}, or deal with conversations such as \cite{li-etal-2017-dailydialog}, which are outside the focus of this work. 

\textbf{COVID-19 Survey Data}\footnote{Data available at \texttt{https://github.com/ben-aaron188/\\covid19worry}} was presented in \cite{kleinberg2020measuring}. This data was collected via a survey in the UK in April 2020, when the UK was under lockdown. The responses of 2500 participants included a text response as well as demographic data (e.g. gender) and emotion-related ratings entered by the participants themselves. We focused on the ``chosen emotion'': a category which the participants chose out of several emotion options. Besides this attribute, each record also had a rating from 1 to 9 for several emotions (anger, fear, worry, etc). The dataset is mainly focused on worry and anxiety due to the topic of the survey. We kept the emotions that were representing at least 4\% of the total records resulting in a dataset of 2408 records.

\textbf{GoEmotions Data}\footnote{Data available at \texttt{https://github.com/google-research/\\google-research/tree/master/goemotions}} was presented in \cite{demszky2020goemotions}. The original dataset contains about 58 thousand Reddit comments with human annotations mapped to 27 emotions or neutral. For our work, we removed neutral comments and kept records that were assigned only a single emotion label (as we are focusing on single label classification) that is in the Ekman taxonomy\cite{ekman1992argument}: Anger, Disgust, Fear, Joy, Sadness, and Surprise. The reason for keeping only Ekman taxonomy emotions was to align this dataset with the next dataset, WASSA-21 (see next item). This dataset already comes with distinct sets for train and test. We trained on the train set (4343 records after removing neutral and non-single emotion records) and tested on the test set (553 records).

\textbf{WASSA-21 Data}\footnote{Data available at \texttt{https://competitions.codalab.org/\\competitions/28713}} was part of the WASSA\footnote{WASSA stands for the 11th Workshop on Computational Approaches to Subjectivity, Sentiment \& Social Media Analysis} 2021 Shared Task on Empathy Detection and Emotion Classification summarized in \cite{tafreshi2021wassa}. This dataset is an extension of \cite{buechel-etal-2018-modeling}'s dataset based on news articles related to harm to an individual, group, nature, etc. The dataset contains essays in which authors expressed their empathy and distress in reactions to these news articles. The essays are annotated for empathy and distress, as well as personality traits and demographic information (age, gender, etc.). Each essay is also tagged with one of the Ekman’s emotions \cite{ekman1992argument}: Anger, Disgust, Fear, Joy, Sadness, and Surprise. We only focused on the emotion for each essay, not the empathy or distress labels. The WASSA-21 dataset already comes with distinct sets for train, dev (development), and test. We trained on the training set (1585 records) and tested on the dev set (245 records). Both datasets are heavily dominated by Sadness (about 40\% of the records) and Anger (22\% in train and 31\% in test). 

Table \ref{tab-data} shows a comparison of the datasets: the emotions and their distribution in each dataset, as well as the total number of records and the average length in characters of the records in each dataset. As shown in the Table, the datasets differ in the emotion categories as well as the distribution: COVID-19 Survey is heavily skewed towards anxiety (57\%), which does not exist in the Ekman taxonomy \cite{ekman1992argument} followed by the other two datasets. GoEmotions and WASSA-21 are both dominated by anger and sadness, although Goemotions is more balanced overall than the other two datasets. The GoEmotions dataset is about double in number of records compared to the other two datasets, but has much shorter text in each record than the other two datasets. In essense, GoEmotions has sentence-length records, while the other two datasets contain essays made up of multiple sentences.

\begin{table}[t]
\centering
\caption{Dataset Comparison: Emotions, their distribution, number of records and average length of a textual record in the Datasets. A ``$-$'' indicates the emotion is not in the Dataset.}
\label{tab-data}
\begin{tabular}{lccc}
\hline
\textbf{Emotion}        & \textbf{COVID-19 Survey} & \textbf{GoEmotions} & \textbf{WASSA-21}    \\
\hline
~~Anger          & 4\%             & 24\%       & 22\% / 31\% \\
~~Anxiety        & 57\%            & $-$          & $-$           \\
~~Disgust        & $-$ & 12\%       & 5\% / 9\%   \\
~~Fear           & 10\%            & 10\%       & 13\%        \\
~~Joy            & $-$               & 20\%       & 6\%         \\
~~Relaxation     & 14\%            & $-$          & $-$           \\
~~Sadness        & 15\%            & 19\%       & 40\%        \\
~~Surprise       & $-$               & 17\%       & 6\% / 10\%  \\
\hline
\textbf{$n$ (train / test) }& 2408            & 4343 / 553 & 1585 / 245 \\
\hline
\textbf{Avg length }& 633  & 66 & 443 \\
\hline
\end{tabular}
\end{table}

Finally, for our experiments, we turned text in all datasets to lower case and removed stop words using NLTK\footnote{\texttt{https://www.nltk.org/}}. Both WASSA-21 and GoEmotions sets have predefined training and test/development sets as described above. For the COVID-19 Survey Dataset, we used an 80-20 split and report the average of the results of 5 runs (standard deviation was below 2\%).

\section{Methodology}
\label{sec-method}
\subsection{Models}
In this work, we explored traditional techniques as well as current state-of-the-art Deep Learning. The models we used are briefly described in the next paragraphs. We used Google colab\footnote{\texttt{https://colab.research.google.com/}} to run all our experiments.

\textbf{Bag-of-Words (BoW).} BoW techniques do not take the textual order of the words into account and instead rely on a word-to-document matrix which contains frequency counts: how often each word is found in the text records. Besides a numeric count of word frequency, we also used TfIdf (Term Frequency–Inverse Document Frequency), which is a statistical measure used to evaluate the importance of a word in a document in a corpus. Using  TfIdf, the importance increases proportionally to the frequency of the word in the document, but it is offset by the frequency of the word in the corpus. We experimented with the usual models such as Naive Bayes, Support Vector Machines, Linear Regression, etc. We used the defaults for all models in scikit-learn\footnote{\texttt{https://scikit-learn.org}} \cite{scikit-learn} to run our BoW experiments.

\textbf{Transformed-Based Deep Learning.} Among the more recent neural architectures, transformer-based models \cite{vaswani-attent} are considered state-of-the-art in several NLP tasks. A Transformer combines a multi-head self-attention mechanism with an encoder-decoder. BERT (Bidirectional Encoder Representations from Transformers) is a transformer-based language model developed by Google \cite{devlin2019bert}. BERT's architecture is based on a multi-layer bidirectional Transformer. In our work, we also explored the use of two extensions of BERT: RoBERTa (Robustly optimized BERT approach) \cite{liu2019roberta} which removed and modified some parts of BERT and trained with more data to make it more robust; and ELECTRA \cite{clark2020electra} which has a different pre-training approach from BERT. We chose these specific models (as opposed to other transformer-based models or even more traditional deep learning such as CNNs or LSTMs) because they were used successfully in recent papers and shared tasks such as the WASSA-21 \cite{tafreshi2021wassa} and performed well in own early experiments.

For these experiments, we used Pytorch and HuggingFace\footnote{\texttt{https://huggingface.co/transformers}}. 
Specifically, we experimented with the \texttt{bert-uncased} model, the \texttt{roberta-base model}, the \texttt{electra-small-discriminator} model and the \texttt{electra-large-discriminator} model.  
For all models, we reported the results for 5 Epochs, learning rate of $1e^-5$, \textit{maxlen} of 256 and batch of 8. We experimented with other values and these seemed to be the best overall in performance.

\subsection{Metrics} 
We report our results based on the classification metrics defined below:

\begin{equation}
Precision ~(Recall) = \frac{TP}{TP+FP ~(FN)}
\end{equation}
\begin{equation}
Accuracy = \frac{TP+TN}{N}
\end{equation}
\begin{equation}
F1\textnormal{-}score=\frac{2 \times Precision \times Recall}{Precision+Recall}
\end{equation} 

where $TP$ is True Positives, $FP$ is False Positives, $FN$ is False
Negatives, and $N$ is the total number of records.

Besides Accuracy, we chose to also report the F1-macro which averages the F1-score over the classes: the
macro-averaged F1 is better suited for showing algorithm effectiveness on smaller categories (see \cite{altrabsheh2014sentiment}), which is important as we are working with imbalanced datasets.

\section{Experimental Results}
\label{sec-results}
Table \ref{table-results-regular} shows the results for the three datasets we used in this paper, only showing the top three transformer-based models and the top two BoW models for each dataset (ordered by f1-macro). As seen in Table \ref{table-results-regular}, the transformer-based models performed better than the BoW models for all three datasets, as expected. The RoBERTa model performed the best in each case. Among the datasets, GoEmotions results were much higher than for the other two datasets. The f1-macro results for GoEmotions were in the mid 70's for the BoW models to low 80's for the transformer-based models, while the highest f1-macro was 49\% for COVID-19 Survey and 54\% for WASSA-21. 

These results make sense as the GoEmotions dataset has double the records, follows a more balanced  distribution of emotion categories, and includes short text records (sentence-length comments taken from reddit). In contrast, the WASSA-21 and the COVID-19 datasets are more highly skewed towards one or two emotions (anxiety for COVID-19 Survey and anger/sadness for WASSA-21). Also, both datasets contain essay-type records with several sentences which sometimes contain multiple emotions. For example, a record in the COVID-19 survey dataset may have had ``chosen emotion'' as anxiety, but the participant also gave a high rating to fear and sadness. We only used ``chosen emotion'' from this data for our experiments, and this shows how different emotions may overlap in the same essay-type text. This was also discussed in the original paper for the WASSA-21 data \cite{tafreshi2021wassa} where the authors observed that when an essay was misclassified in their experiments, the essay often contained many emotions.

\begin{table}[t]
\centering
\caption{Results for each dataset, ordered by f1-macro}
\begin{tabular}{llcc}
\hline
\textbf{DataSet} & \textbf{Model} & \textbf{Accuracy} & \textbf{f1-macro} \\ \hline
COVID-19 Survey & RoBERTa & 64\%  & 49\%  \\
& BERT & 58\% & 45\% \\
& ELECTRA-large & 63\% & 34\%\\
& Linear SVM  &  59\%              & 23\%                 \\ 
& Linear Regression       & 58\%              & 19\%              \\ \hline

GoEmotions & RoBERTa           & 83\%              & 83\%              \\ 
 & BERT              & 83\%              & 83\%              \\ 
& ELECTRA-small     & 82\%              & 81\%              \\ 
& Linear Regression & 76\%              & 77\%              \\ 
& Linear SVM        & 76\%              & 76\%              \\ 
\hline
WASSA-21 & RoBERTa        & 62\%              & 54\%              \\ 
 & BERT           & 59\%              & 48\%              \\ 
& ELECTRA-Large        & 66\%              & 46\%              \\ 
& Linear SVM     & 53\%              & 29\%              \\ 
& Naive Bayes    & 51\%              & 25\%              \\ 
\hline

\end{tabular}
\label{table-results-regular}
\end{table}

\begin{figure}[t] 	
         \centering
\includegraphics[scale=0.45]{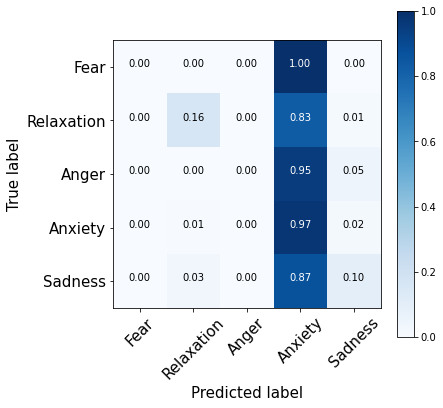}
\caption{Normalized Confusion Matrix of Linear SVM on the COVID-19 Survey Data} 	
\label{fig:fig-conf-svm} 
\end{figure}

\begin{figure}[t]
         \centering
\includegraphics[scale=0.45]{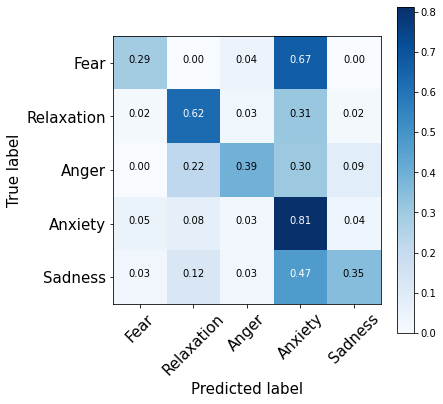}
\caption{Normalized Confusion Matrix of RoBERTa on the COVID-19 Survey Data } 	
\label{fig:fig-conf-roberta} 
\end{figure}

In addition, we explored the results by inspecting the confusion matrix of specific runs. Given the COVID-19 Survey data, an example result for the Linear SVM confusion matrix is shown in Figure \ref{fig:fig-conf-svm} and for RoBERTa in Figure \ref{fig:fig-conf-roberta}. Both figures show a normalized confusion matrix (per the number of records in each class). Figure \ref{fig:fig-conf-svm} shows that the Linear SVM classified most records as the dominant class, anxiety (57\% of the entire data, see Table \ref{tab-data}). In contrast, RoBERTa classified the different emotion categories overall much better as seen in Figure \ref{fig:fig-conf-roberta}, though it still assigns many records from other emotions to the dominant class, anxiety.

Machine Learning models, and especially Deep Learning models, are known to perform better when they are trained on larger sets of records. For the datasets that did not perform as well, we used data augmentation to explore possible improvements in performance. 

For the WASSA-21 data, we used a total of 5928 records for training: we augmented the WASSA-21 Train Set with the data from the GoEmotions Train Set. As both sets (original WASSA-21 and our preprocessed GoEmotions - see Section \ref{sec-data}) follow the Ekman taxonomy \cite{ekman1992argument}, this augmentation was straightforward. The results are shown in Table \ref{table-results-wassa-goemo}. The models that showed improvement are the RoBERTa and the linear SVM, while the rest of the models stay at same levels of accuracy and f1-macro.

\begin{table}[t]
\centering
\caption{Results of training with (WASSA-21 Train Set + GoEmotions Train Set) and Testing on the WASSA-21 Dev Set. Bold numbers indicate improvement over original model, $\Delta$ is the difference from the original f1-macro in Table \ref{table-results-regular}.}
\begin{tabular}{lccc}
\hline
\textbf{Model} & \textbf{Accuracy} & \textbf{f1-macro} & $\Delta$          \\ 
\hline
RoBERTa        & \textbf{67\%}           & \textbf{60\%} & 6\%\\ 
BERT           & 57\%              & 48\%                  &  0\%\\
ELECTRA-Large        & \textbf{64\%}              & 46\%         & 0\%     \\
Linear SVM     & \textbf{55\%}            & \textbf{38\%} &  9\%    \\
Naive Bayes    & 51\%              & 25\%          &0\%    \\ \hline
\end{tabular}
\label{table-results-wassa-goemo}
\end{table}

\begin{table}[t]
\centering
\caption{Results of training with (COVID-19 Train Set + GoEmotions Train Set) and Testing on the COVID-19 Test Set. Bold numbers indicate improvement over original model, $\Delta$ is the difference from the original f1-macro in Table \ref{table-results-regular}.}
\begin{tabular}{lccc}
\hline
\textbf{Model} & \textbf{Accuracy} & \textbf{f1-macro} & $\Delta$          \\ 
\hline

RoBERTa           & 60\%    & \textbf{51\%}             & 2\%              \\
BERT              &\textbf{63\%}   & \textbf{48\%}        & 3\%                \\
ELECTRA-Large     & \textbf{65\%}     & \textbf{43\%}           & 9\%                  \\
Linear SVM        & \textbf{60\%}     & \textbf{30\%}     & 7\% \\
Linear Regression & 57\%              & \textbf{28\%}     & 9\%    \\     \hline

\end{tabular}
\label{table-results-covid-goemo}
\end{table}

For the COVID-19 Survey data, we used GoEmotions train data as well, though the emotion categories are different (see Table \ref{tab-data} for emotion categories and distributions). Therefore, we only added records from the GoEmotion train dataset whose label was anger, fear or sadness. This resulted in a train set of 4198 records. The results are shown in Table \ref{table-results-covid-goemo}. In this case, all results improved for all models and some models show large improvements such as the ELECTRA-large with 9\% improvement of the f1-macro. 

Overall, data augmentation seems to improve the performance of the classifiers by adding more records and also making the dataset more balanced. However, the data in GoEmotions are quite different from the data in the other two datasets (reddit comments versus essays) so the f1-macro results are still not higher than 50's for the COVID-19 Survey data and 60's for the WASSA-21 data.

\subsection{Case Study: Testing example records from social media}
We also explored the applicability of the models to the task of recognizing emotions from actual posts on social media. Due to time limitations, a full study is left for future work; for this paper, we pulled a few posts from reddit, manually annotated them with emotion labels, and then tested two of the models in the previous section. Specifically, the posts were collected from subreddits (forums) `r/Anxiety' and `/r/COVID19\_support' and date in 2021. The models we tested are the top RoBERTA models based on (a) GoEmotion and (b) COVID-19 Survey Data: our justification was that these datasets are either based on reddit (GoEmotions) or related to COVID-19 topic (COVID-19 Survey). In the following, we include part of the reddit comment, the emotion label for the emotion assigned by us, and the resulting emotion classification label for (a) and for (b).

\begin{enumerate}
\item (r/Anxiety) ``\textit{I tested positive for Covid back in [...] and it was like my whole world collapsed. Instantly as i found out i started to shake and chain smoked [...] The weight on my shoulders was unbearable. [...] With constant watching of the news and social media built up a storm i was unaware about within my body}''. Our label: `Fear', (a) predicted `Fear', (b) predicted `Anxiety'.

\item (r/Anxiety)``\textit{I just want to say how appreciative I am for each and every one of you. Reading these comments has calmed me down completely and brought me back to reality. [...] hearing it from regular people like me makes it a million times more powerful}''. Our label: `Joy', (a) predicted `Joy', (b) predicted `Sadness'. 

\item (r/COVID-19 Support) ``\textit{This is one of my main problems during the pandemic! I hate missing out on all the big events because it seems like most of my friends are still getting together. It’s hard}''. Our label: `Sadness', (a) predicted `Sadness', (b) predicted `Sadness'.

\item  (r/COVID-19 Support) ``\textit{That is SO incredibly frustrating [...] was probably ignorantly walking around maskless thinking everyone was fine with it. when in reality no one wanted to be the one to reprimand a superior. Ridiculous}''. Our label: `Anger', (a) predicted `Anger', (b) predicted `Anger'.

\item (r/COVID-19 Support) ``\textit{I can't wait to get my final shot [...] imma go back to the gym!!!! gained so much weight and people keep pointing it out}''. Our label: `Joy', (a) predicted `Joy', (b) predicted `Sadness'. 

\item (r/COVID-19 Support) ``\textit{My daughter was born the night before they declared a pandemic. I went into the hospital [...], and the world had been turned upside down [...] basically been a nightmare since}''. Our label: `Fear', (a) predicted `Sadness', (b) predicted `Fear'. 
\end{enumerate}

From this small experiment, we see that the (a) model which is based on GoEmotions data does better, and the (b) model based on COVID-19 Survey data assigned `Sadness' to `Joy' labeled records. The `Joy' emotion does not exist in the COVID-19 Survey data, but the model did not assign the closest emotion either, which is `Relaxation'. 

\section{Conclusions}
\label{sec-conclusion}
In this paper, we presented an extensive experimentation with three different datasets that were recently presented to the research community. The datasets on which we focused exhibit a variety of characteristics: size in number of records, average length of text in the records, emotion taxonomies and distributions, self ratings versus human annotated labels, among others. Our experiments involved a variety of models for emotion classification. We also explored data augmentation as a means to improve performance. Overall, RoBERTa was the highest performing model and data augmentation did work for most cases. We also tested the applicability of the best models on example records we pulled from social media posts and labeled ourselves with emotions. Our discussion highlights the complexity of emotion prediction or classification using text. 

Future directions include expanding our research to other datasets and additional models, exploring the use of lexicons such as the NRC EmoLex \cite{Mohammad13} as well as additional characteristics from the data (e.g. author demographics), and experimenting with multi-label classification and sentence-level classification as a means to improve essay-level classification. 

\bibliographystyle{./IEEEtran}
\bibliography{refs}

\vspace{12pt}
\end{document}